\documentclass[letterpaper]{article}

\usepackage{arxiv}

\usepackage[utf8]{inputenc} 
\usepackage[T1]{fontenc}    
\usepackage{hyperref}       
\usepackage{url}            
\usepackage{booktabs}       
\usepackage{amsfonts}       
\usepackage{amsmath}
\usepackage{nicefrac}       
\usepackage{microtype}      
\usepackage{lipsum}		
\usepackage{graphicx}
\usepackage{natbib}
\usepackage{doi}
\usepackage{bm}

\title{Alternative positional encoding functions for neural transformers}

\author{ \href{https://orcid.org/0000-0001-8231-5687}{\includegraphics[scale=0.06]{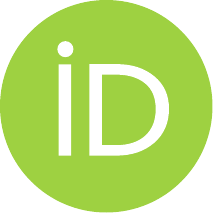}\hspace{1mm}Ezequiel L\'opez-Rubio}\thanks{Corresponding author: Ezequiel L\'opez-Rubio. Ezequiel L\'opez-Rubio and Rafael Marcos Luque-Baena are also with ITIS Software. Universidad de M\'alaga. C/ Arquitecto Francisco Peñalosa 18, 29010, Málaga, Spain} \\
	Department of Computer Languages \\ and Computer Science\\
    University of M\'alaga\\
    Bulevar Louis Pasteur, 35\\
    29071 M\'alaga, Spain \\
	\texttt{ezeqlr@lcc.uma.es} \\
	\And
	\href{https://orcid.org/0009-0005-3275-9614}{\includegraphics[scale=0.06]{orcid.pdf}\hspace{1mm}Macor\'is Decena-Gim\'enez} \\
	Department of Systems Engineering \\ and Automation\\
	University of M\'alaga\\
    Bulevar Louis Pasteur, 35\\
    29071 M\'alaga, Spain \\
	\texttt{macorisd@uma.es} \\
	\And
	\href{https://orcid.org/0000-0001-5536-1805}{\includegraphics[scale=0.06]{orcid.pdf}\hspace{1mm}Rafael Marcos Luque-Baena} \\
	Department of Computer Languages \\ and Computer Science\\
    University of M\'alaga\\
    Bulevar Louis Pasteur, 35\\
    29071 M\'alaga, Spain \\
	\texttt{rmluque@uma.es} \\
}

\renewcommand{\shorttitle}{Alternative positional encoding functions}

\hypersetup{
pdftitle={Alternative positional encoding functions for neural transformers},
pdfsubject={cs.AI, cs.CL},
pdfauthor={Ezequiel L\'opez-Rubio, Macor\'is Decena-Gim\'enez, Rafael Marcos Luque-Baena},
pdfkeywords={neural transformers, positional encoding, language models, periodic functions},
}

\begin{document}
\maketitle

\begin{abstract}
	A key module in neural transformer-based deep architectures is positional encoding. This module enables a suitable way to encode positional information as input for transformer neural layers. This success has been rooted in the use of sinusoidal functions of various frequencies, in order to capture recurrent patterns of differing typical periods. In this work, an alternative set of periodic functions is proposed for positional encoding. These functions preserve some key properties of sinusoidal ones, while they depart from them in fundamental ways. Some tentative experiments are reported, where the original sinusoidal version is substantially outperformed. This strongly suggests that the alternative functions may have a wider use in other transformer architectures.
\end{abstract}


\section{Introduction\label{sec:Introduction}}

Transformer architectures are now the dominant models for sequence processing in natural language and other modalities, yet the underlying self-attention operation is permutation-invariant and thus has no intrinsic notion of order \citep{dufter2022position}.  To make Transformers sensitive to word order or temporal structure, positional encoding (PE) mechanisms inject information about token positions, either as absolute indices or as relative distances between tokens \citep{dufter2022position}.  The design of positional encoding has emerged as a central inductive bias that strongly affects performance, robustness, and length generalization \citep{dufter2022position,kazemnejad2023impact}.

A standard Transformer layer applies content-based self-attention followed by position-wise feed-forward networks, which means that, without additional signals, reordering the input tokens leaves the output unchanged \citep{dufter2022position,vaswani2017attention}.  Position information can be introduced at the input level, within attention matrices, or before the output, corresponding broadly to position embeddings, attention manipulation, and hybrid schemes \citep{dufter2022position,shaw2018self}.  An extensive survey shows that these approaches can be clustered into absolute and relative methods, with further distinctions by injection point and functional form \citep{dufter2022position,kazemnejad2023impact,chi2022kerple}.

Absolute positional encodings assign each sequence index a dedicated vector that is combined with token embeddings, for instance via elementwise addition \citep{dufter2022position,vaswani2017attention}.  The original Transformer uses fixed sinusoidal encodings, while many subsequent variants employ learned absolute embeddings whose parameters are optimized along with token embeddings \citep{dufter2022position}.  Absolute schemes are simple and effective on moderate sequence lengths but tend to generalize poorly when models are evaluated on contexts significantly longer than those seen during training \citep{kazemnejad2023impact}.

Relative positional encodings instead represent the distance between token pairs and inject this information directly into the attention computation \citep{dufter2022position,shaw2018self}.  In these models, attention logits are augmented with terms that depend on the relative offset of query and key positions, allowing the model to learn distance-sensitive patterns that are invariant under global shifts of the sequence \citep{dufter2022position,shaw2018self}.  This perspective unifies a large family of methods, including those that use learned relative embeddings or structured functions of the distance \citep{dufter2022position,chi2022kerple}.

A prominent recent line of work develops kernelized relative positional embeddings for length extrapolation \citep{chi2022kerple}.  In this framework, distances between positions are mapped through conditionally positive definite kernels, which are then incorporated into the attention scores in a way that preserves the probabilistic interpretation of self-attention \citep{chi2022kerple}.  Empirical results show that appropriate kernel choices, such as logarithmic variants, can yield strong extrapolation to much longer sequences than those used during training, often outperforming standard relative schemes on language modeling benchmarks \citep{chi2022kerple}.

\section{Methodology\label{sec:Methodology}}

Next, the proposed alternative positional encoding functions are detailed.
Transformer architectures rely on positional encodings to inject order
information into sequences processed by permutation-invariant self-attention
layers \citep{vaswani2017attention}.

The original Transformer uses a deterministic periodic encoding that
assigns to each position $m\in\{0,\dots,L-1\}$ a vector $\mathrm{PE}(m)\in\mathbb{R}^{d_{\text{model}}}$
\citep{vaswani2017attention,kazemnejad2019transformerPosEncoding}:
\begin{align}
\mathrm{PE}(m,2i) & =\varphi\left(\frac{m}{10000^{2i/d_{\text{model}}}}\right)\label{eq:sinusoidal-even}\\
\mathrm{PE}(m,2i+1) & =\psi\left(\frac{m}{10000^{2i/d_{\text{model}}}}\right)\label{eq:sinusoidal-odd}
\end{align}
 where $0\le i<d_{\text{model}}/2$. The standard choice for the periodic
functions is sinusoidal, i.e., $\varphi=\sin$, $\psi=\cos$.

The encoding is then added to the token embeddings $\bm{x}_{m}$,
\begin{equation}
\tilde{\bm{x}}_{m}=\bm{x}_{m}+\mathrm{PE}(m),
\end{equation}
 before being fed to the first self-attention layer. The choice of
exponentially spaced frequencies allows the model to represent relative
offsets as approximately linear functions of the encodings and to
extrapolate to longer sequences.

Rotary Positional Embedding (RoPE) encodes positions by rotating query
and key vectors in a shared complex (or 2D) subspace, so that their
inner product depends on relative position \citep{su2021roformer,eleuther2024rope}.
Consider a per-head query $\bm{q}_{m}\in\mathbb{R}^{d_{k}}$ and key
$\bm{k}_{n}\in\mathbb{R}^{d_{k}}$ at positions $m$ and $n$. Split
each into $d_{k}/2$ two-dimensional components and define a rotation
for each pair: 
\begin{equation}
R_{\theta}\left(m\right)=\begin{bmatrix}\psi\left(m\theta\right) & -\varphi\left(m\theta\right)\\
\varphi\left(m\theta\right) & \psi\left(m\theta\right)
\end{bmatrix}
\end{equation}

Again, the standard choice for the periodic functions is sinusoidal,
i.e., $\varphi=\sin$, $\psi=\cos$.

RoPE applies $R_{\theta_{j}}(m)$ and $R_{\theta_{j}}(n)$ to the
$j$-th 2D component of $\bm{q}_{m}$ and $\bm{k}_{n}$, respectively:
\begin{align}
\bm{q}'_{m} & =R\left(m\right)\bm{q}_{m},\\
\bm{k}'_{n} & =R\left(n\right)\bm{k}_{n},
\end{align}
 where $R(m)$ is block-diagonal with 2D rotations along the diagonal.
The attention is then computed using the rotated vectors: 
\begin{equation}
e_{mn}=\frac{\bm{q}'_{m}\cdot\bm{k}'_{n}}{\sqrt{d_{k}}}.
\end{equation}
 Because $\bm{q}'_{m}\cdot\bm{k}'_{n}$ depends on $m-n$ through
the rotation angles, RoPE effectively injects relative position information
while preserving absolute encodings.

Now, we propose to employ non sinusoidal functions for $\varphi$
and $\psi$. Two restrictions are imposed to keep the key features
of the original method:
\begin{enumerate}
\item Periodicity. The functions $\varphi:\mathbb{R}\rightarrow\mathbb{R}$
and $\psi:\mathbb{R}\rightarrow\mathbb{R}$ must be periodic real
valued functions with period $\left[0,2\pi\right]$. 
\item Phase shift. The functions $\varphi$ and $\psi$ must have the same
shape but different phase:\\
\begin{equation}
\psi\left(m\right)=\varphi\left(\frac{\pi}{2}-m\right)\label{eq:PhaseShift}
\end{equation}
\end{enumerate}
Given (\ref{eq:PhaseShift}), only the function $\varphi$ must be
specified, because $\psi$ is readily obtained.

The periodic continuous piecewise linear function $\varphi=\mathrm{tri}$,
the square wave function $\varphi=\mathrm{sqw}$, and the sawtooth
function $\varphi=\mathrm{saw}$ are proposed, where:

\begin{equation}
\mathrm{tri}\left(m\right)=\begin{cases}
\frac{2m}{\pi} & \textrm{if }m\in\left[0,\frac{\pi}{2}\right]\\
-\frac{2}{\pi}m+2 & \textrm{if }m\in\left[\frac{\pi}{2},\frac{3\pi}{2}\right]\\
\frac{2m}{\pi}-4 & \textrm{if }m\in\left[\frac{3\pi}{2},2\pi\right]\\
\mathrm{tri}\left(\mathrm{mod}\left(m,2\pi\right)\right) & \mathrm{otherwise}
\end{cases}\label{eq:tri-function}
\end{equation}

\begin{equation}
\mathrm{sqw}\left(m\right)=\begin{cases}
-1 & \textrm{if }m\in\left[0,\pi\right)\\
1 & \textrm{if }m\in\left[\pi,2\pi\right)\\
\mathrm{sqw}\left(\mathrm{mod}\left(m,2\pi\right)\right) & \mathrm{otherwise}
\end{cases}\label{eq:sqw-function}
\end{equation}

\begin{equation}
\mathrm{saw}\left(m\right)=\begin{cases}
m & \textrm{if }m\in\left[0,\pi\right)\\
m-2\pi & \textrm{if }m\in\left[\pi,2\pi\right)\\
\mathrm{saw}\left(\mathrm{mod}\left(m,2\pi\right)\right) & \mathrm{otherwise}
\end{cases}\label{eq:saw-function}
\end{equation}
where $\mathrm{mod}$ stands for the modulus of the floating point
division. 
\begin{center}
\begin{figure}
\begin{centering}
\includegraphics[width=\textwidth]{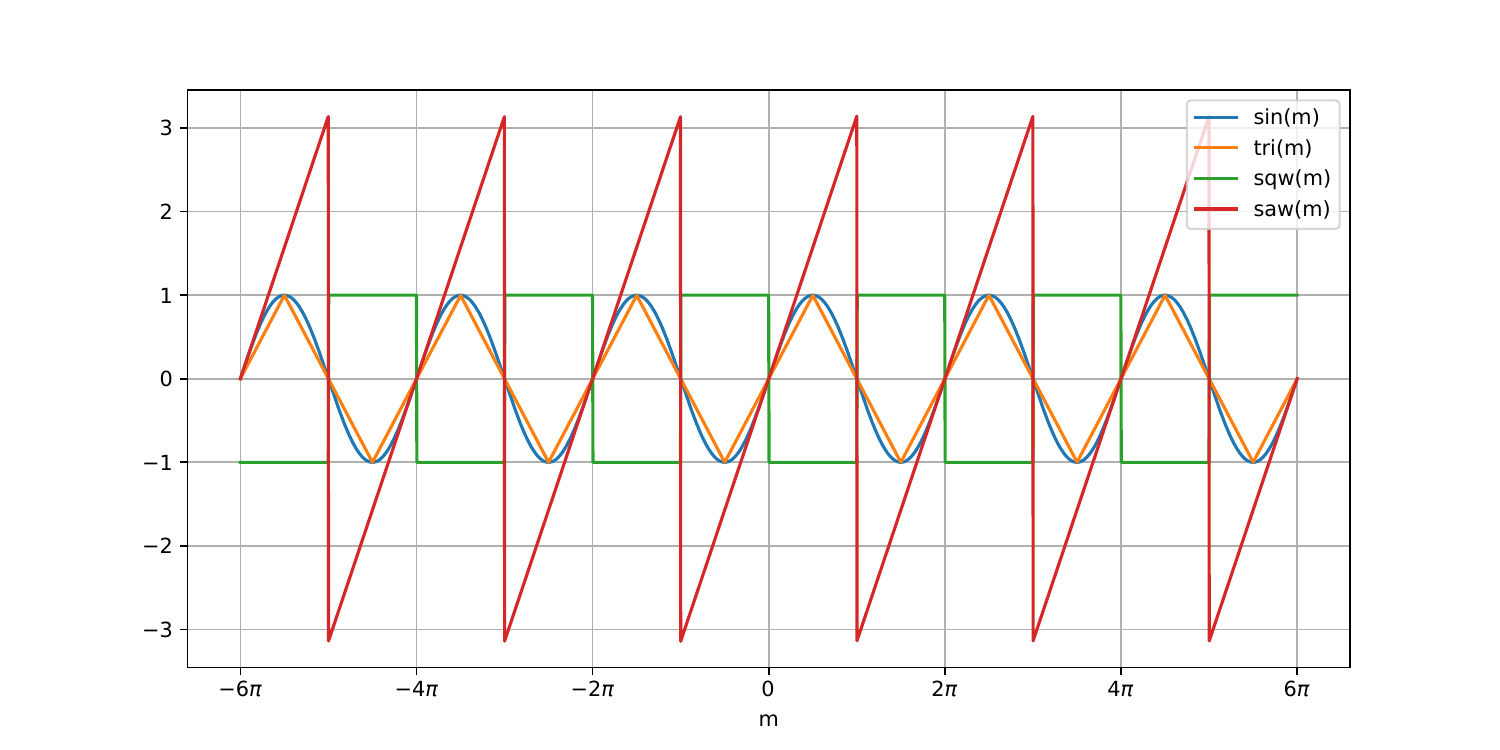}
\par\end{centering}
\caption{\label{fig:PE-functions}The standard sinusoidal function and the
proposed alternative functions.}

\end{figure}
\par\end{center}

\section{Experiments\label{sec:Experiments}}

All experiments were carried out on a single NVIDIA RTX 6000 Ada Generation GPU (48~GB VRAM) running Ubuntu 22.04, Python 3.10, and PyTorch 2.9.1. The code was based on the public implementation of the original Transformer \citep{vaswani2017attention} by \citet{ko2019transformer}. More specifically, we used the fork by \citet{zhao2025translation} that already contained the pre-processed data required for training. Our own fork, available at \url{https://github.com/macorisd/alt-positional-encoding-transformer}, added four interchangeable positional encoding functions:
\begin{enumerate}
    \item \textbf{Sinusoidal}, i.e. the original fixed encoding;
    \item \textbf{Triangular}, equation (\ref{eq:tri-function});
    \item \textbf{Square}, equation (\ref{eq:sqw-function});
    \item \textbf{Sawtooth}, equation (\ref{eq:saw-function}).
\end{enumerate}

The model architecture followed the Transformer base configuration \citep{vaswani2017attention}, with $d_{\text{model}} = 512$, $N=6$ layers in both encoder and decoder, $h = 8$ attention heads, a feed--forward dimension of $d_{\text{ff}} = 2048$, and a dropout probability of 0.1 applied to all sub--layers.

We trained and evaluated the model on the Multi30K English--German image--description dataset \citep{elliott2016multi30k}, which provided parallel English--German captions aligned at the sentence level. All text was lowercased and tokenized at the word level using language--specific tokenizers. Special tokens were added to each sequence, including \texttt{<sos>} and \texttt{<eos>} to mark sentence boundaries, \texttt{<unk>} for out--of--vocabulary words, and \texttt{<pad>} for sequence padding. Separate vocabularies were built for the source (English) and target (German) languages using only the training split, discarding tokens with a frequency lower than two. Sentences were converted to sequences of token indices, with unseen tokens mapped to \texttt{<unk>}. During training, batches were dynamically padded to the maximum sequence length within each batch, with a maximum allowed length of 256 tokens. A batch size of 512 sentence pairs was used for all experiments.

To systematically compare the four positional encoding variants, we employed 10--fold cross--validation using only the Multi30K training split, which consists of 29{,}001 sentence pairs. The training data was randomly shuffled with a fixed seed and partitioned into 10 folds. For each positional encoding variant, we trained 10 models using 9 folds for training and 1 fold for validation, rotating the held-out fold across runs. Models were trained using the Adam optimizer \citep{kingma2014adam} with an initial learning rate of $10^{-5}$ and weight decay of $5 \times 10^{-4}$. The learning rate was automatically decayed based on validation performance. Gradients were clipped to a maximum norm of 1.0, and training used cross-entropy loss with padding tokens ignored.

For each encoding function, we report the final training/validation loss and BLEU-4 after the last epoch, as well as the best validation BLEU-4 observed during training. Table~\ref{tab:summary-results} summarizes the mean performance across the 10 cross-validation folds.

\begin{table}[htbp]
    \centering
    \caption{Average performance over 10-fold cross-validation (1000 epochs per fold). Values are reported as mean $\pm$ standard deviation, and the best results are highlighted in bold.}
    \label{tab:summary-results}
    \small
    \begin{tabular}{lcccc}
        \toprule
        \textbf{Encoding Function} & \multicolumn{2}{c}{\textbf{Loss}} &
                         \multicolumn{2}{c}{\textbf{BLEU-4}}\\
        \cmidrule(lr){2-3}\cmidrule(lr){4-5}
         & Final Train & Final Val & Final  & Best \\ 
        \midrule
        Sinusoidal   & $3.05 \pm 0.03$ & $3.12 \pm 0.03$ & $29.48 \pm 0.76$ & $29.63 \pm 0.77$\\
        Triangular   & $\mathbf{2.41 \pm 0.01}$ & $2.57 \pm 0.02$ & $40.68 \pm 0.36$ & $40.78 \pm 0.37$\\
        Square       & $2.64 \pm 0.07$ & $2.74 \pm 0.06$ & $34.54 \pm 1.54$ & $34.93 \pm 1.72$\\
        Sawtooth     & $\mathbf{2.41 \pm 0.08}$ & $\mathbf{2.53 \pm 0.10}$ & $\mathbf{40.77 \pm 2.65}$ & $\mathbf{41.03 \pm 2.60}$ \\
        \bottomrule
    \end{tabular}
\end{table}

\begin{figure}[t]
  \centering
  \includegraphics[width=\textwidth]{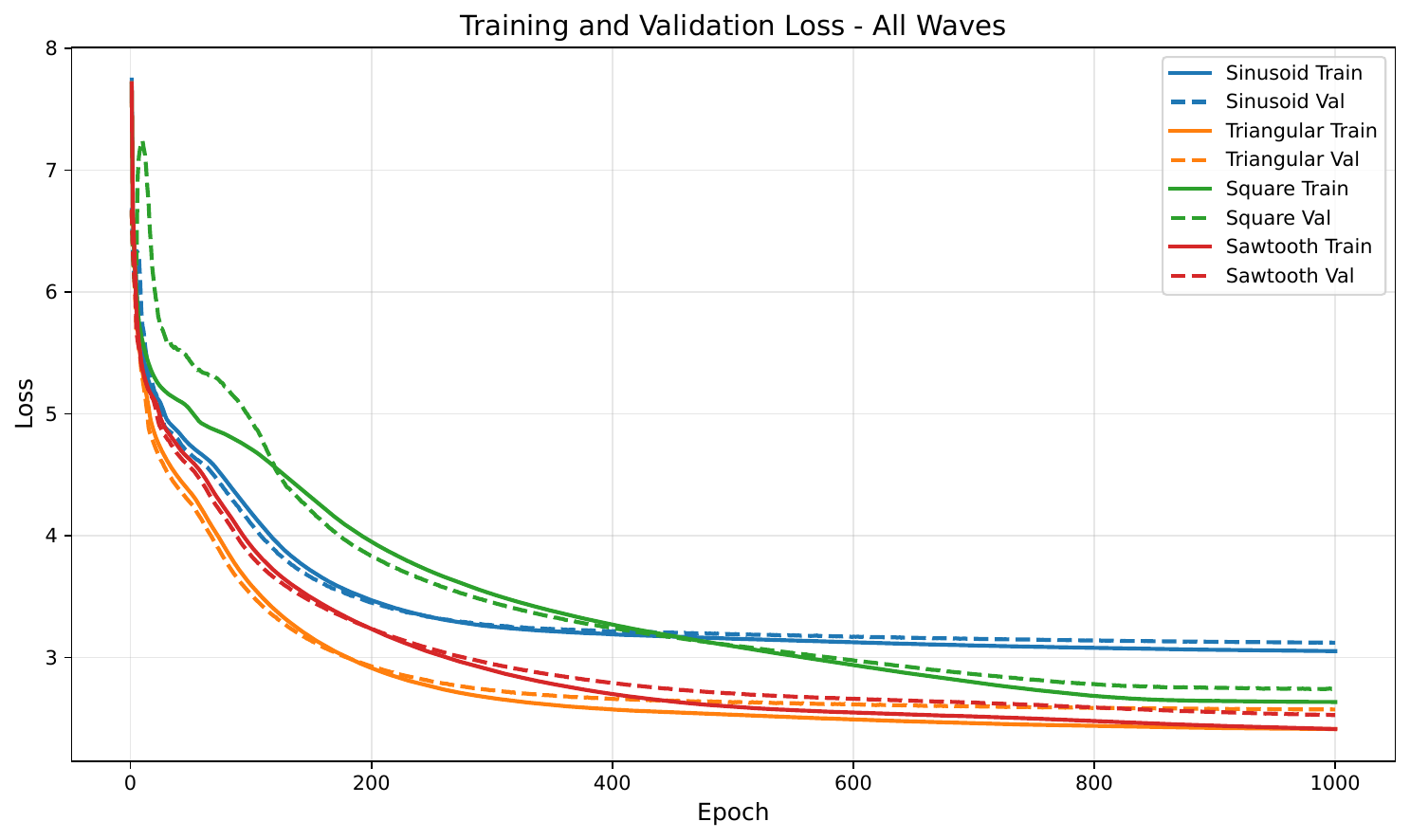}
  \caption{Training dynamics of the loss across the four positional encoding variants. The plot shows the average training loss (solid lines) and validation loss (dashed lines) as a function of the training epoch. All curves are averaged over the 10 cross-validation folds.}
  \label{fig:loss-graph}
\end{figure}

\begin{figure}[t]
  \centering
  \includegraphics[width=\textwidth]{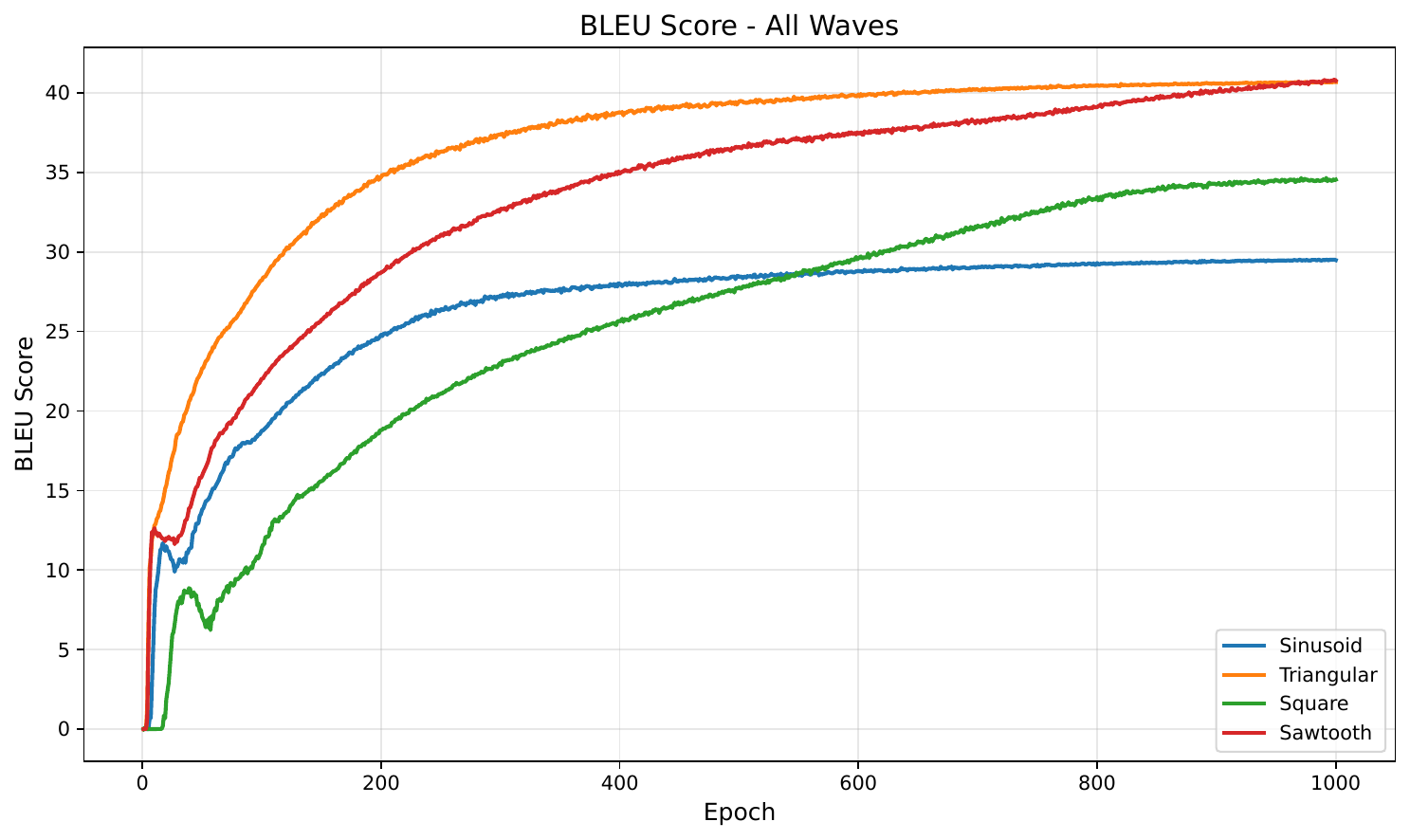}
  \caption{Training dynamics of validation BLEU-4 across the four positional encoding variants. The plot shows the average validation BLEU-4 score as a function of the training epoch. All curves are averaged over the 10 cross-validation folds.}
  \label{fig:bleu-graph}
\end{figure}

Figures~\ref{fig:loss-graph} and~\ref{fig:bleu-graph} provide a detailed view of the training dynamics underlying the summary statistics in Table~\ref{tab:summary-results}. Figure~\ref{fig:loss-graph} shows the evolution of the average training loss and validation loss across epochs for each positional encoding variant. Figure~\ref{fig:bleu-graph} shows the average validation BLEU-4 score across epochs for the same variants. All curves are averaged over the 10 cross-validation folds.

\section{Discussion\label{sec:Discussion}}

The key features of our proposal and the main implications of the experimental results are discussed next. The sinusoidal functions have been the standard choice for most of the work done on fixed positional encoding for neural transformers. Our present work proposes three alternative periodic functions, each with its completely different features, that can also be employed for all kinds of neural transformers. Therefore, a suitable function can be chosen to fit the problem at hand. In particular, each of the proposed functions exhibits essential features that sinusoidal
functions do not have:
\begin{itemize}
\item The piecewise linear function $\varphi=\mathrm{tri}$, equation (\ref{eq:tri-function}), has a piecewise
constant slope that distributes its outputs more uniformly over the
range of the function, as compared to the sinusoidal function that
compresses the output values associated with input values close to
integer multiples of $\frac{\pi}{2}$ into smaller intervals of the
output range.
\item The square wave function $\varphi=\mathrm{sqw}$, equation (\ref{eq:sqw-function}),  quantizes the input
values into a discrete set of possible output values. This way, the
function is a quantizer of its input domain.
\item The sawtooth function $\varphi=\mathrm{saw}$, equation (\ref{eq:saw-function}), has the same slope for
all points where it is differentiable. Moreover it distributes its
ouputs uniformly over its range, like the piecewise linear function.
\end{itemize}

Given the wide range of applications of neural transformers, it is envisaged that each function may be the best performer for a particular set of problems.

The results of the experiments reported in Section \ref{sec:Experiments} give some relevant information. For the benchmark problem considered, all three alternative functions clearly outperform the standard sinusoidal function, both in cross-entropy loss and BLEU score terms. The evolution of the learning process is stable in all cases, reaching a steady state. The best performing functions are triangular and sawtooth, although the triangular function is faster to learn. The number of epochs that the learning process takes to reach the steady state is critical since it is directly related to the energy consumption of such a process. Therefore, it may be more advantageous to choose a fast learning function (the triangular one in this case), rather than the best performer (the sawtooth function).

Future work includes more experimentation with a wider range of problems so that the preliminary results presented here are further validated. Other alternative periodic functions may also be considered.

\section{Conclusions\label{sec:Conclusions}}
A new set of functions for the positional encoding module of neural transformers has been proposed. These functions differ from the standard sinusoidal functions in several relevant ways. The proposal has been tested on a well-known benchmark task. It has been found that the alternative functions clearly surpass the original sinusoidal approach. These results open the path for more successful applications of these encoding functions. In particular, significant performance enhancements and energy savings may be obtained.

\section*{Author contributions}
Conceptualization, E.L.-R.; methodology, E.L.-R.; software, M.D.-G.; validation, E.L.-R. and R.M.L.-B.; formal analysis, E.L.-R.; investigation, E.L.-R.; resources, M.D.-G.; data curation, M.D.-G.; writing---original draft preparation, E.L.-R. and M.D.-G.; writing---review and editing, E.L.-R., M.D.-G. and R.M.L.-B.; visualization, M.D.-G.; supervision, E.L.-R.; project administration, E.L.-R.; funding acquisition, E.L.-R. and R.M.L.-B. All authors have read and agreed to the published version of the manuscript.

\bibliographystyle{unsrtnat}
\bibliography{references}  






\end{document}